\documentclass[10pt,twocolumn,letterpaper]{article}

\usepackage{cvpr}              % To produce the CAMERA-READY version
\usepackage{pifont}
\newcommand{\cmark}{\ding{51}}%
\newcommand{\xmark}{\ding{55}}%
\usepackage{booktabs}   % for \toprule, \midrule, \bottomrule, \cmidrule
\usepackage{multirow}   % for multirow cells (used for the dataset grouping)
\usepackage{graphicx}   % for \resizebox
\usepackage{amssymb}    % for \checkmark and \xmark symbols
\usepackage{pifont}     % for \xmark (\ding{55}) if not already defined
\usepackage{bm}
\usepackage{booktabs} % for nice rules
\usepackage{tabularx} % for full-width tables
\usepackage{pifont}   % for \xmark
\usepackage{amssymb}  % for \checkmark
\usepackage{array}    % for \newcolumntype
\usepackage{float}
\usepackage[table]{xcolor} % for coloring cells
\usepackage[accsupp]{axessibility}
\newcommand{\goodcell}[1]{\cellcolor{green!30}{#1}}
\newcommand{\badcell}[1]{\cellcolor{red!30}{#1}}
\newcolumntype{C}{>{\centering\arraybackslash}X} % centered stretchable column

\newcommand{\pf}[1]{\cellcolor{red!30}#1}
\newcommand{\ps}[1]{\cellcolor{orange!30}#1}

\definecolor{cvprblue}{rgb}{0.21,0.49,0.74}
\usepackage[pagebackref,breaklinks,colorlinks,allcolors=cvprblue]{hyperref}

\def\confName{CVPR}
\def\confYear{2026}

\title{Generalizable Sparse-View 3D Reconstruction from Unconstrained Images}
\author{
Vinayak Gupta$^{1}$ \quad
Chih-Hao Lin$^{2}$ \quad
Shenlong Wang$^{2}$ \quad
Anand Bhattad$^{3}$ \quad
Jia-Bin Huang$^{1}$\\[0.45em]
\small $^{1}$University of Maryland, College Park \quad
\small $^{2}$University of Illinois Urbana-Champaign \quad 
\small $^{3}$Johns Hopkins University\\[0.55em]
\href{https://genwildsplat.github.io/}{\Large{\tt\color{cvprblue}https://genwildsplat.github.io/}}
}
\begin{document}
\twocolumn[{%
\renewcommand\twocolumn[1][]{#1}%
\maketitle\vspace{-2em}
\centering
\includegraphics[width=\textwidth]{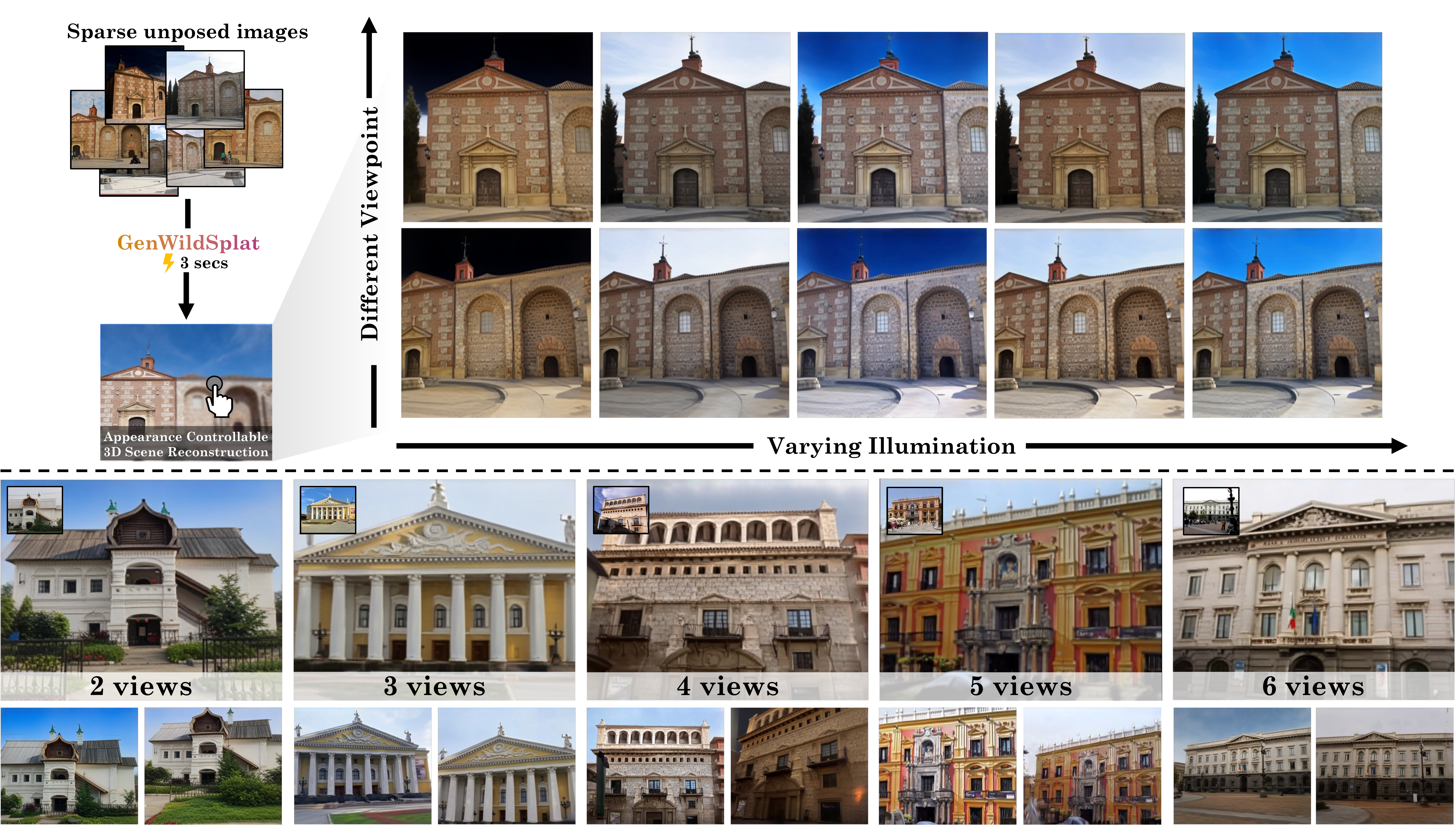}
\vspace{-6mm}
\captionof{figure}{
\textbf{GenWildSplat} reconstructs 3D scenes from sparse, unposed images with varying illumination and transient objects in a single 3-second feed-forward pass, and no per-scene optimization is required. Given 2–6 input views, our method predicts novel views under target lighting conditions while handling occlusions. \textbf{Top:} Novel-view synthesis under different lighting from the same sparse inputs, demonstrating appearance control. \textbf{Bottom:} Reconstruction quality across varying input sparsity (2–6 views), showing view-consistent rendering even with minimal observations. In each block, the top-left image (inset) is an input view; the remaining images are novel-view predictions under novel lighting. All scenes are unseen during training, demonstrating strong generalization to real-world environments.
}
\vspace{4mm}
\label{fig:teaser}
}]

% \begin{figure*}[t!]
%     \centering
%     \includegraphics[width=\linewidth]{fig/Teaser.pdf}
%     \caption{
% }
%     \label{fig:placeholder}
    
% \end{figure*}

\begin{abstract}
Reconstructing 3D scenes from sparse, unposed images remains challenging under real-world conditions with varying illumination and transient occlusions. Existing methods rely on scene-specific optimization using appearance embeddings or dynamic masks, which requires extensive per-scene training and fails under sparse views. Moreover, evaluations on limited scenes raise questions about generalization. 
We present \textbf{GenWildSplat}, a feed-forward framework for sparse-view outdoor reconstruction that requires no per-scene optimization. 
Given unposed internet images, GenWildSplat predicts depth, camera parameters, and 3D Gaussians in a canonical space using learned geometric priors. 
An appearance adapter modulates appearance for target lighting conditions, while semantic segmentation handles transient objects. Through curriculum learning on synthetic and real data, GenWildSplat generalizes across diverse illumination and occlusion patterns. 
Evaluations on PhotoTourism and MegaScenes benchmark demonstrate state-of-the-art feed-forward rendering quality, achieving real-time inference without test-time optimization.

\end{abstract}

\section{Introduction}

Reconstructing 3D scenes from 2D images is crucial for applications such as AR/VR and navigation~\cite{mur2015orb, zhou2024comprehensive}. 
Extending these techniques to in-the-wild imagery remains challenging due to three factors: 
(1) Internet photos show wide lighting variations across time and seasons, 
(2) handheld captures contain transient occluders like tourists or vehicles that must be excluded, and 
(3) real-world scenes often provide sparse viewpoints, unlike curated multi-view datasets. 
Effective reconstruction requires disentangling static scene content from dynamic lighting and transient objects. 
Prior NeRF~\cite{chen2022hallucinated, rudnev2022nerf, sun2022neural} and Gaussian Splatting~\cite{kulhanek2024wildgaussians, xu2024wild, dahmani2024swag, wang2024we, kaleta2025lumigauss, wang2025look} methods rely on \emph{per-scene optimization} and dense views, while sparse-view in-the-wild approaches are time-intensive. 
Feed-forward models~\cite{wang2025vggt, jiang2025anysplat, xu2025depthsplat} enable real-time reconstruction but are limited to fixed lighting and fail under dynamic conditions. In Tab.~\ref{tab:nerf_comparison}, we show the key characteristic differences between the previous methods.
While these methods perform well on benchmarks like PhotoTourism~\cite{snavely2006photo}, they fail on more challenging datasets such as MegaScenes~\cite{tung2024megascenes} (Fig.~\ref{fig:failure}), which feature sparse views, diverse lighting, and heavy occlusions.

To address these limitations, we introduce \textbf{GenWildSplat}, a generalizable model for fast feed-forward 3D scene reconstruction from sparse in-the-wild scenes without requiring per-scene optimization. To our knowledge, this is the first approach to integrate both appearance and occlusion modeling within a feed-forward 3D reconstruction paradigm. The key insight is that combining large-scale synthetic and real-world data enables the model to learn robust associations across diverse illumination conditions. Our framework uses VGGT’s transformer to process sparse, unposed multi-view images into rich feature maps, which specialized heads decode into depth, camera parameters, and per-pixel Gaussians. The resulting set of attributes defines a canonical representation that captures a unified scene geometry disentangled from illumination. 
However, directly decoding these canonical Gaussians into a novel view via a differentiable rasterizer leads to multi-view inconsistencies.

To effectively map the canonical space to a desired target lighting, we introduce an appearance adapter that conditions on the target lighting and transforms the canonical Gaussian colors to the corresponding target lighting space. 
We parameterize lighting information using a light code estimated by a light encoder, which represents it in a compact latent space. 
To handle transient objects, we leverage a pre-trained segmentation network that identifies dynamic elements, such as people or cars. 
It produces explicit occlusion masks that guide our model to ignore transient regions during supervision, ensuring a clean and stable 3D scene. 
Ideally, one would train the model using multi-view images under varying illumination to render novel views under new lighting conditions. However, the absence of such multi-view, multi-lighting datasets makes direct supervised learning infeasible.

% We adopt a training strategy that requires no paired multi-illumination data and can operate on any unordered set of input images. 
% Each image is encoded into a compact light code representing scene illumination, and the appearance adapter conditions on this code, along with the canonical Gaussian colors, to produce transformed colors. These transformed colors are then rasterized to reconstruct the original image and provide supervision through reconstruction error. Direct training on real-world datasets~\cite{tung2024megascenes}, however, often leads to collapse or poor rendering, as learning geometry and illumination simultaneously in sparse-view settings is highly ill-posed. To address this, GenWildSplat is trained using a curriculum strategy, first on synthetic data to learn appearance, then progressively incorporating occlusions, and finally fine-tuned on real-world MegaScenes data, enabling stable learning and effective generalization.
We train GenWildSplat without paired multi-view multi-illumination data, using unordered image collections. Each input image is mapped to a compact light code, and the appearance adapter conditions on this code and the canonical Gaussian colors to generate transformed colors, which are supervised via image reconstruction. Direct training on large-scale real-world data is unstable, as jointly learning geometry and illumination from sparse views is a highly ill-posed problem. To avoid collapse, we adopt a curriculum: first, learn appearance on synthetic data; and finally, add synthetic occlusions. This staged strategy enables stable optimization and strong generalization.

\begin{table}[t!]
    \centering
    \caption{Comparison of key characteristics across existing methods and our approach. Unlike optimization-based or feed-forward baselines, our method is fast, capable of sparse views, view-consistent, and generalizes to novel lighting conditions.}
    \label{tab:nerf_comparison}
    \scriptsize
    \resizebox{\linewidth}{!}{
    \begin{tabular}{lccccc}
        \toprule
        \textbf{Method} & \textbf{In-the-Wild} & \textbf{Fast} & \textbf{View-Consis} & \textbf{Few-Views}  \\
        \midrule
        Optimization-Based~\cite{tang2024nexussplats, kulhanek2024wildgaussians} & \goodcell{\cmark} & \badcell{\xmark} & \goodcell{\cmark} & \badcell{\xmark} \\
        Feed-Forward Based~\cite{jiang2025anysplat} & \badcell{\xmark} & \goodcell{\cmark} &  \badcell{\xmark} & \goodcell{\cmark} \\
        Ours & \goodcell{\cmark} & \goodcell{\cmark} & \goodcell{\cmark} & \goodcell{\cmark} \\
        \bottomrule
    \end{tabular}
    }
\end{table}

\begin{figure}[t!]
    \centering
    \includegraphics[width=\linewidth]{fig/Failure.pdf}
    \vspace{-7mm}
    \caption{
    \textbf{Limitations of Prior Work.} Prior methods~\cite{kulhanek2024wildgaussians, tang2024nexussplats} fail under sparse-view conditions. \textbf{(a) Overfitting:} Scene-specific optimization produces artifacts and geometric spikes with small camera perturbations. \textbf{(b) Camera dependency:} Methods rely on COLMAP for pose estimation, which fails under sparsity. Even with higher-quality transformer-based poses (e.g., VGGT), reconstructions exhibit severe artifacts and blurring. \textbf{(c) Limited appearance adaptation:} Test-time optimization cannot adapt to novel lighting, causing color bleeding and geometric distortions when target illumination differs from training conditions.
    %\textbf{Limitations of Prior Works}. Prior methods~\cite{kulhanek2024wildgaussians, tang2024nexussplats} struggle to reconstruct scenes from sparse views. (a) Scene-specific models often overfit to the training images, producing artifacts and spikes even with small camera shifts. (b) Prior methods rely on COLMAP-estimated cameras, but COLMAP fails under sparse views, but even when supplied with higher-quality transformer-based predictions such as VGGT, these approaches still exhibit substantial artifacts and blurring. (c) These methods also require test-time optimization and hence, when the target lighting differs significantly from training conditions, they fail to adapt, causing colour bleeding and geometric distortions.
    }
    \label{fig:failure}
    \vspace{-10pt}
\end{figure}

\begin{figure*}
    \centering
    \includegraphics[width=\linewidth]{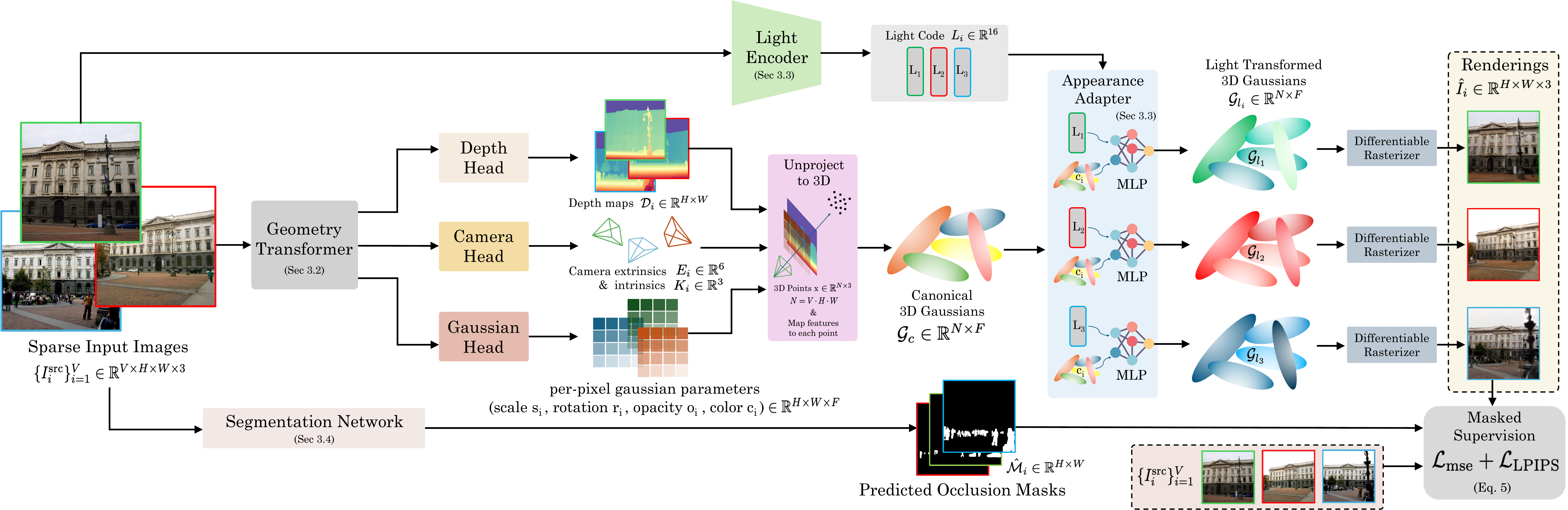}
    \\\vspace{-5pt}
    \caption{
    \textbf{Overview of GenWildSplat.} Given sparse, unposed images $\{I_i\}_{i=1}^V$ with appearance variations and transient objects, a geometry transformer extracts multi-view features $\mathbf{F}_i$ encoding semantic and geometric information. Specialized prediction heads process these features to output per-pixel depth $\mathbf{D}_i$, camera parameters $(\mathbf{K}_i, \mathbf{E}_i)$, and Gaussian attributes, which are unprojected into canonical 3D Gaussians $\mathbf{G}_c$. A light encoder $\mathcal{E}_{Light}$ extracts per-image lighting codes $\mathbf{L}_i = \mathcal{E}_{Light}(I_i)$. An MLP $F_{\text{light}}$ modulates the canonical Gaussian colors using these codes: $\mathbf{G}_{\ell_i} = F_{\text{light}}(\mathbf{G}_c, \mathbf{L}_i)$. Each set of transformed Gaussians $\mathbf{G}_{\ell_i}$ is rasterized to reconstruct $\hat{I}_i$. A pre-trained segmentation network provides occlusion masks $M_i$ to identify transient objects. Masked reconstruction loss focuses supervision on static content, enabling photorealistic, view-consistent reconstruction from sparse in-the-wild imagery.
}
    \label{fig:method}
        \vspace{-5pt}
\end{figure*}

To evaluate generalization on unseen scenes, we benchmark GenWildSplat on both the Phototourism and the more challenging MegaScenes datasets. Our method consistently outperforms existing approaches~\cite{zhang2024gaussian, kulhanek2024wildgaussians, tang2024nexussplats} in reconstructing accurate scene geometry, modeling appearance under varying illumination, and effectively handling occlusions. 
Interestingly, GenWildSplat surpasses scene-specific methods~\cite{kulhanek2024wildgaussians, tang2024nexussplats} that rely on per-image appearance optimization and test-time fine-tuning, demonstrating the strength of leveraging large-scale pre-trained priors for robust 3D understanding. 
Overall, GenWildSplat represents a step toward generalizable 3D reconstruction, offering a feed-forward, illumination- and occlusion-aware framework that scales to diverse, real-world environments for real-time 3D scene understanding.

\label{sec:intro}

\setlength{\tabcolsep}{5pt} % adjust horizontal spacing

\section{Related Works}
\label{sec:formatting}

\textbf{Optimization‐based Novel View Synthesis (NVS)} reconstructs 3D scenes from 2D images for novel viewpoint generation. \textit{Gaussian Splatting (3DGS)}~\cite{kerbl20233d} represents scenes with explicit 3D Gaussian primitives, enabling real‐time rasterization via a CUDA pipeline. Extensions improve depth and camera regularization for few‐view settings~\cite{zhu2024fsgs, xiong2025sparsegs, paliwal2024coherentgs, yin2024fewviewgs}, but per‐scene optimization is still required, limiting fast, test‑time use.  

\noindent \textbf{Feed‑forward Novel View Synthesis (NVS)} predicts 3D Gaussians without scene‑specific tuning, either assuming known poses (\textit{pose‑aware}) or estimating poses during inference (\textit{pose‑free}).  

\underline{\textit{Pose‑aware}} methods use calibrated poses and include: (1) direct 3D Gaussian predictors~\cite{charatan2024pixelsplat, chen2024mvsplat, chen2024mvsplat360, wang2024freesplat, xu2025depthsplat}; (2) transformer‑based LRM decoders~\cite{lrm, gslrm, xu2024grm, ziwen2024long}; and (3) latent feed‑forward models~\cite{jinlvsm}. 
These are fast but rely on accurate camera poses.  

\underline{\textit{Pose-free}} methods jointly estimate poses and novel views, with DUSt3R~\cite{wang2024dust3r} and MASt3R~\cite{mast3r} predicting depth and fusing dense 3D. Subsequent works~\cite{wang20243d, liu2025slam3r, murai2025mast3r, wang2025continuous, wang2025vggt, yang2025fast3r, tang2025mv} extend this using transformer cascades for unified pose, trajectory, and geometry estimation, while latent approaches~\cite{jiang2025rayzer} self-supervise pose and view prediction. Despite strong performance, these methods degrade under lighting variation or dynamic distractors.

\noindent \textbf{Novel View Synthesis in the Wild} reconstructs 3D scenes from unconstrained photo collections, challenged by (\textit{i}) varying illumination and (\textit{ii}) transient objects.  

\underline{\textit{Varying Appearance}} is handled via per‑view latent embeddings~\cite{martin2021nerf, yang2023cross}, CNN‑conditioned features~\cite{zhang2024gaussian, wang2024we}, hash‑grid fields~\cite{dahmani2024swag}, or hierarchical light decoupling~\cite{tang2024nexussplats}. 
Most require long test‑time optimization (10+ hours). Diffusion-based strategies have also been explored to harmonize illumination across views~\cite{alzayer2025generative} or to enable training-free multi-view consistent editing~\cite{alzayer2025coupled}.

\underline{\textit{Occlusion Modeling}} addresses moving objects via robust regression~\cite{sabour2023robustnerf}, uncertainty features~\cite{ren2024nerf, kulhanek2024wildgaussians}, 2D occlusion masks~\cite{zhang2024gaussian}, or per‑image and per‑Gaussian transient embeddings~\cite{tang2024nexussplats}. 
These methods remain slow, require intensive training, and result from a lack of 3D priors.

\noindent In contrast, our feed‑forward approach directly processes sparse unposed images under varying lighting and dynamics, reconstructing 3D scenes with controllable appearance and view‑consistent rendering in 3 seconds, without per‑scene optimization.

\begin{figure*}
    \centering
    \includegraphics[width=\linewidth]{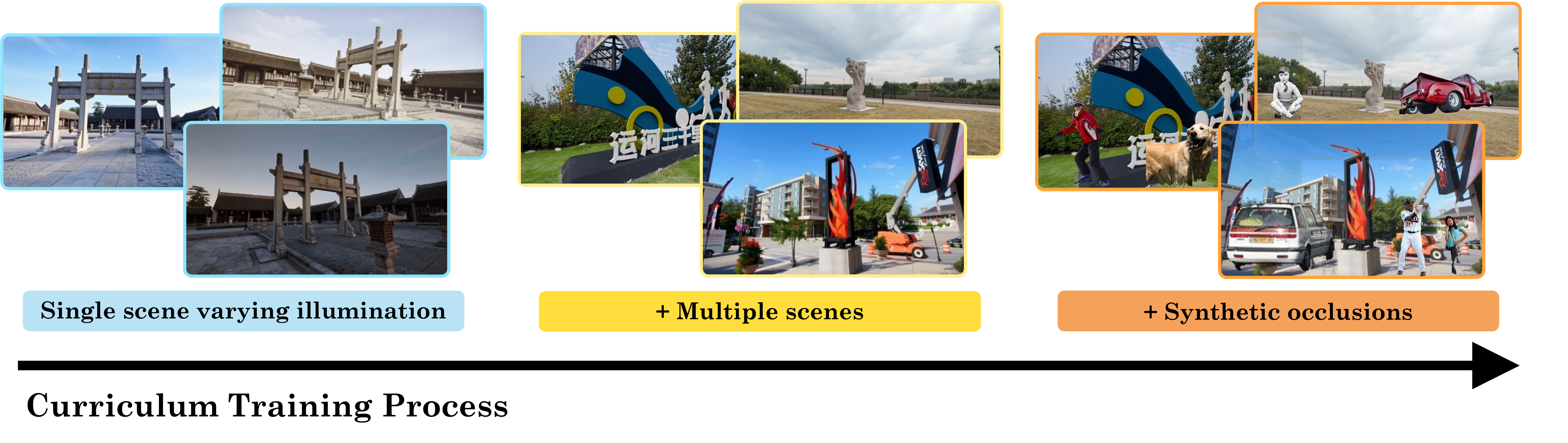}
    \vspace{-10pt}
    \caption{
    \textbf{Curriculum Learning.} 
    Training proceeds in three stages. 
    \textbf{Stage I:} Single scene with illumination variation. In this stage, the model learns to disentangle lighting from geometry. 
    \textbf{Stage II:} Multiple scenes: the model then learns geometric and appearance priors across diverse environments. 
    \textbf{Stage III:} Synthetic occlusions: the network learns to handle transient objects and multi-view inconsistencies. 
    Despite training only on synthetic data, the model generalizes to real-world appearance variations and occlusions.
    }
    \label{fig:curriculum}
    \vspace{-5pt}
\end{figure*}

\section{Method}

\subsection{Preliminary: AnySplat}  

Our method builds upon AnySplat~\cite{jiang2025anysplat}, a feed-forward framework that reconstructs 3D scenes as Gaussian primitives from multiple input images in a single pass.

\noindent \textbf{Architecture.} AnySplat takes unposed images and processes them through a VGGT~\cite{wang2025vggt} transformer backbone to extract multi-view features. Three prediction heads then output: (a) depth maps for each view, (b) camera poses and (c) 3D Gaussian properties including position, color, shape, and opacity. To reduce redundancy from per-pixel Gaussian prediction, AnySplat voxelizes the scene, assigns confidence scores, and merges overlapping Gaussians within each voxel to form a compact 3D representation.

\noindent \textbf{Training.} AnySplat trains without ground-truth 3D data. Instead, it uses VGGT's pretrained model to generate pseudo-labels for depth and camera poses. The predicted Gaussians are rendered back to 2D and supervised against the input views, ensuring the 3D representation remains consistent with the observed images.

\subsection{Problem Formulation and Overview}
Given unposed input images 
\(\mathcal{I} = \{I_1, I_2, \dots, I_N\}\) 
captured under varying illumination with transient objects, we reconstruct a 3D scene that renders novel views under different appearance conditions while handling occlusions.

Our model predicts 3D Gaussians conditioned on target appearance \(\mathbf{L}\) as $\boldsymbol{\mathcal{G}_l} = f_\theta(\mathcal{I}, \mathbf{L})$, where each Gaussian \(g_{L} \in \boldsymbol{\mathcal{G}_l}\) is parameterized by:
\[
g_l = \{\boldsymbol{\mu}, \boldsymbol{\sigma}, \boldsymbol{r}, \boldsymbol{s}, \boldsymbol{c}_L\},
\]
with position \(\boldsymbol{\mu} \in \mathbb{R}^3\), opacity \(\boldsymbol{\sigma} \in \mathbb{R}^+\), rotation \(\boldsymbol{r} \in \mathbb{R}^4\), scale \(\boldsymbol{s} \in \mathbb{R}^3\), and appearance-dependent spherical harmonic (SH) coefficients \(\boldsymbol{c}_L \in \mathbb{R}^{75}\). 
%(25 coefficients × 3 color channels).

\noindent\textbf{Architecture.} 
A VGGT transformer backbone \(\phi_\theta\) extracts multi-view features \(\boldsymbol{F} = \phi_\theta(\mathcal{I})\). Similar to Anysplat, three prediction heads process these features:
\[
\boldsymbol{D} = h_D(\boldsymbol{F}), \, (\boldsymbol{K}, \boldsymbol{E}) = h_C(\boldsymbol{F}), \, (\boldsymbol{s}, \boldsymbol{r}, \boldsymbol{\sigma}, \boldsymbol{c}) = h_{\mathrm{gauss}}(\boldsymbol{F}),
\]
where \(h_D\) predicts per-view depth maps \(\boldsymbol{D}\), \(h_C\) estimates camera intrinsics \(\boldsymbol{K}\) and extrinsics \(\boldsymbol{E}\), and \(h_{\mathrm{gauss}}\) outputs appearance-independent Gaussian properties and canonical SH coefficients \(\boldsymbol{c} \in \mathbb{R}^{75}\).
An appearance adapter \(\psi_\theta\) modulates the canonical colors for target appearance:
\begin{equation}
\boldsymbol{c_{L_i}} = \psi_\theta(\boldsymbol{c}, \mathbf{L}_i),
\end{equation}
where \(\mathbf{L}_i \in \mathbb{R}^{d}\) is a learned appearance embedding.

\noindent\textbf{Training.}
Gaussians \(\boldsymbol{\mathcal{G}_l}\) are rasterized via diff. splatting:
\begin{equation}
\hat{I}_j = \mathcal{R}(\boldsymbol{\mathcal{G}_l}, \boldsymbol{K}_j, \boldsymbol{E}_j),
\end{equation}
and trained end-to-end with reconstruction loss. Though trained only on input views, the model generalizes to novel appearance conditions. We adopt a curriculum training strategy to sequentially refine geometry, appearance, and occlusion modeling for stable convergence. We describe our methodology in Fig.~\ref{fig:method}

\begin{figure*}
    \centering
    \includegraphics[width=\linewidth]{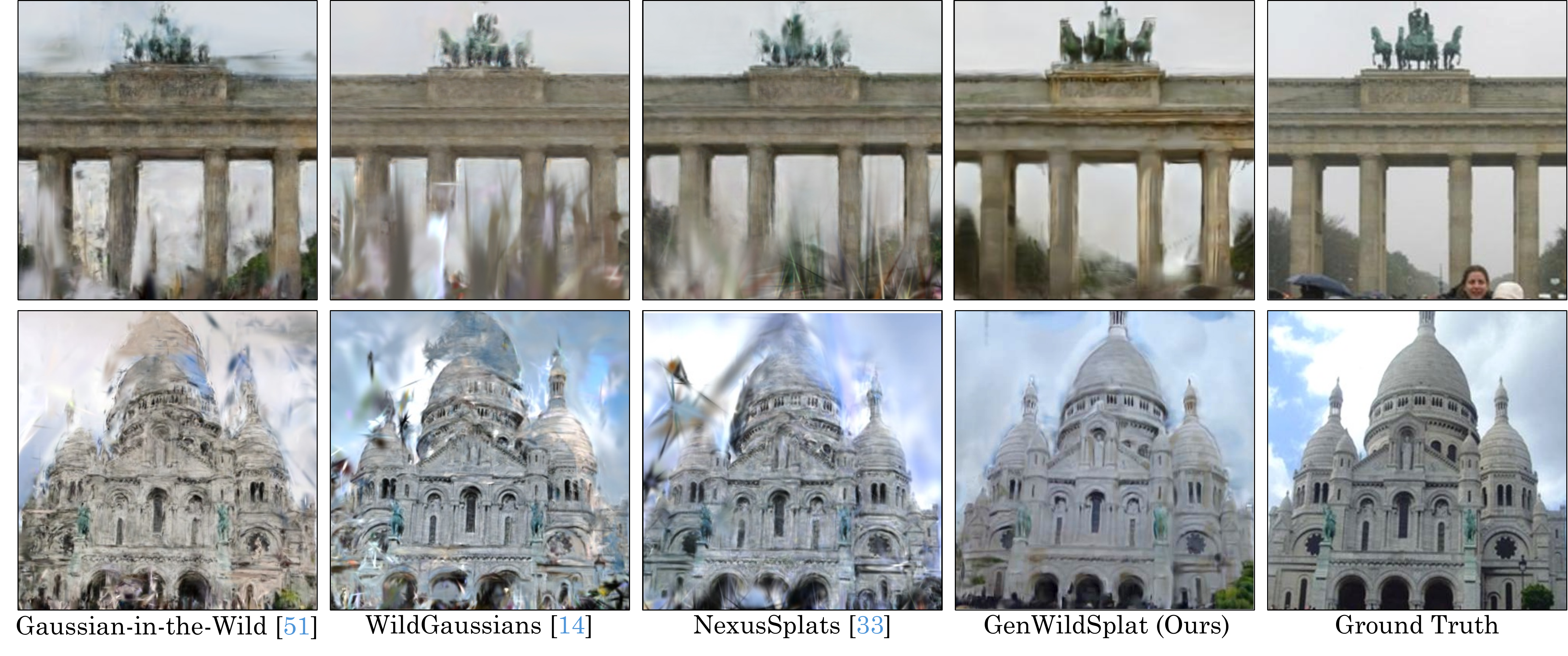}
    \\\vspace{-10pt}
\caption{\textbf{Comparison on the Photo-Tourism dataset against optimization-based methods.} 
Optimization-based methods trained from scratch often struggle to accurately reconstruct scenes from sparse views, even when test-time optimization is applied.
In contrast, our feedforward approach efficiently generates plausible geometry and controllable appearance for complex scenes. As shown in Fig.~\ref{fig:failure}, replacing COLMAP poses with VGGT poses improves their performance; thus, we adopt this modification across all evaluations, thereby solely benefiting the baseline's performance.
    }
    \label{fig:phototourism-sota}
    \vspace{-10pt}
\end{figure*}

\subsection{Appearance Modelling: Appearance Adapter}

Existing methods like WildGaussians~\cite{kulhanek2024wildgaussians} and NexusSplats~\cite{tang2024nexussplats} model appearance using randomly initialized embeddings jointly optimized with geometry during training. 
At test time, these methods require optimizing a new embedding for each novel view or lighting condition, precluding feed-forward inference. We instead predict all scene parameters, including appearance, in a single forward pass.

Our \textit{Appearance Adapter} transforms Gaussian colors to match a target lighting condition. A 2D CNN-based encoder $\mathcal{E}_{Light}$ extracts per-view light codes, which an MLP $F_{\text{light}}$ uses to modulate the Gaussian colors $\mathcal{G}_c = [\mathbf{c}_1, \dots, \mathbf{c}_N]^\top \in \mathbb{R}^{N \times 75}$:
\begin{equation}
    \mathbf{L}_i = \mathcal{E}_{Light}(I^{(i)}), \quad i = 1, \dots, V.
\end{equation}
\vspace{-5mm}
\begin{equation}
    \mathcal{G}_{l_i} = F_{\text{light}}\!\left(\mathcal{G}_c, \mathbf{L}_i\right), \quad i = 1, \dots, V,
\end{equation}

where $\mathcal{G}_{l_i} = [\tilde{\mathbf{c}}_{i,1}, \dots, \tilde{\mathbf{c}}_{i,N}]^\top$ are the transformed colors under view $i$'s lighting. Each set of transformed Gaussians is independently rasterized to reconstruct its corresponding input view, enabling self-supervised training without test-time optimization.

\subsection{Occlusion Modelling}
Transient objects (such as people and vehicles) cause floating artifacts and unstable gradients when treated as static geometry. Prior work~\cite{zhang2024gaussian} uses internally predicted visibility maps or uncertainty estimates~\cite{kulhanek2024wildgaussians,oquab2023dinov2}, which can collapse during unsupervised training by down-weighting difficult regions. This incorrectly suppresses static structures, such as trees, that appear in sparse views (Fig.~\ref{fig:failure}).

We instead use a pre-trained semantic segmentation network to detect common transient classes (person, car, bus, truck). These predictions yield a binary mask \(S \in \{0,1\}^{H \times W}\) where \(S(p)=1\) indicates transients. We apply visibility weighting \(M = 1 - S\) directly to images: \(I_{\text{m}} = I \odot M\) and \(\hat{I}_{\text{m}} = \hat{I} \odot M\), focusing on static regions:

\begin{align}
\label{eq:loss_masked}
    \mathcal{L} &=  \text{MSE}(I_{\text{m}}, \hat{I}_{\text{m}}) + \lambda \cdot \text{Percep}(I_{\text{m}}, \hat{I}_{\text{m}})
\end{align}

where \(\hat{I}\) is the rendered image, \(I_{\text{gt}}\) the ground truth, and \(\odot\) denotes elementwise multiplication. Using an external segmentation prior in this way prevents the model from ``explaining away'' transient content by collapsing its own visibility estimate, stabilizes gradients in dynamic regions, and preserves the static structure during training.

\subsection{Curriculum Learning for Large-Scale Training}

Feed-forward reconstruction on unconstrained imagery requires training on large, diverse datasets. Direct training on data with appearance variation and transient objects is unstable, as learning geometry, lighting, and occlusion jointly is difficult. Training only on curated datasets, however, fails to build priors for in-the-wild generalization. We use curriculum learning to break the task into progressive stages (Fig.~\ref{fig:curriculum}), thereby improving convergence and reconstruction quality compared to end-to-end training.

\noindent\textbf{Stage 1: Lighting (Appearance).} Train on a single synthetic scene with illumination variation but no transients, learning lighting or broadly appearance representation without geometric or occlusion confounds. Empirically, we found that this simplified the appearance decomposition for subsequent training without collapsing.

\noindent\textbf{Stage 2: Multi-scene generalization.} Introduce additional synthetic scenes to improve appearance and geometry modeling across diverse environments.

\noindent\textbf{Stage 3: Occlusion handling.} Add synthetic transients where we have access to ground-truth masks for supervision. We then train the model to predict these occlusion masks alongside geometry and appearance, disentangling transients from static content.

\begin{figure*}
    \centering
    \includegraphics[width=\linewidth]{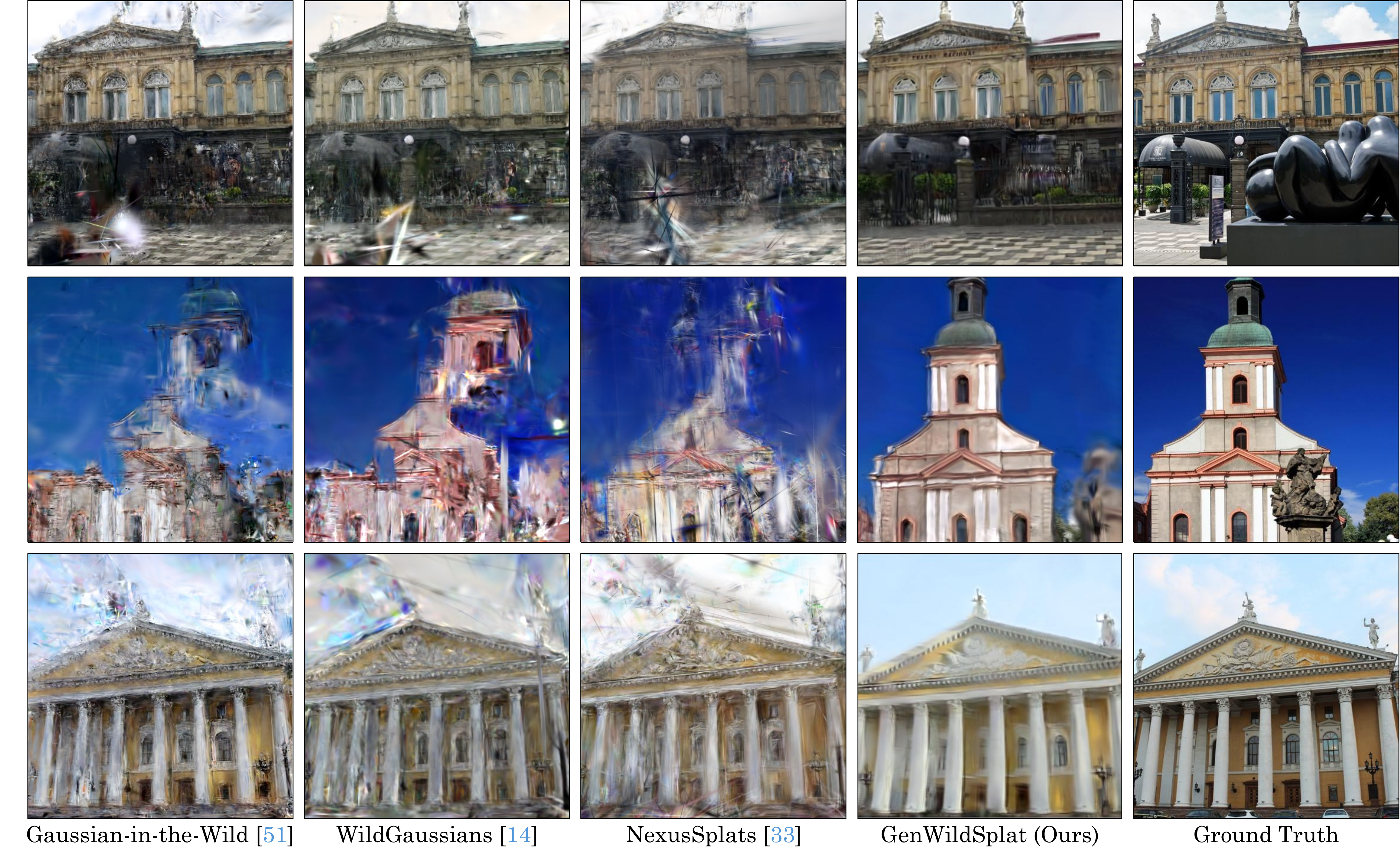}
    \vspace{-7mm}
\caption{\textbf{Comparison on the MegaScenes dataset against optimization-based methods.} 
The MegaScenes dataset poses significant challenges for 3D reconstruction due to wide variations in viewpoints and lighting. Prior SOTA methods often fail, producing artifacts such as noisy ground (row 1), geometric distortions and inconsistencies when rendering novel views (row 2), and spiky/blurred skies (row 3). GenWildSplat, in contrast, generates clean and consistent renderings across diverse scenes, demonstrating robust performance even on these highly challenging in-the-wild settings. }
    \vspace{-3mm}
    \label{fig:megascenes-sota}
\end{figure*}

Despite being trained only on synthetic occlusions and appearance variations, our method generalizes well to real-world sparse-view scenes (Fig.~\ref{fig:phototourism-sota}, Fig.~\ref{fig:megascenes-sota}).

\subsection{Training Framework}

\begin{figure}
    \centering
    \includegraphics[width=\linewidth]{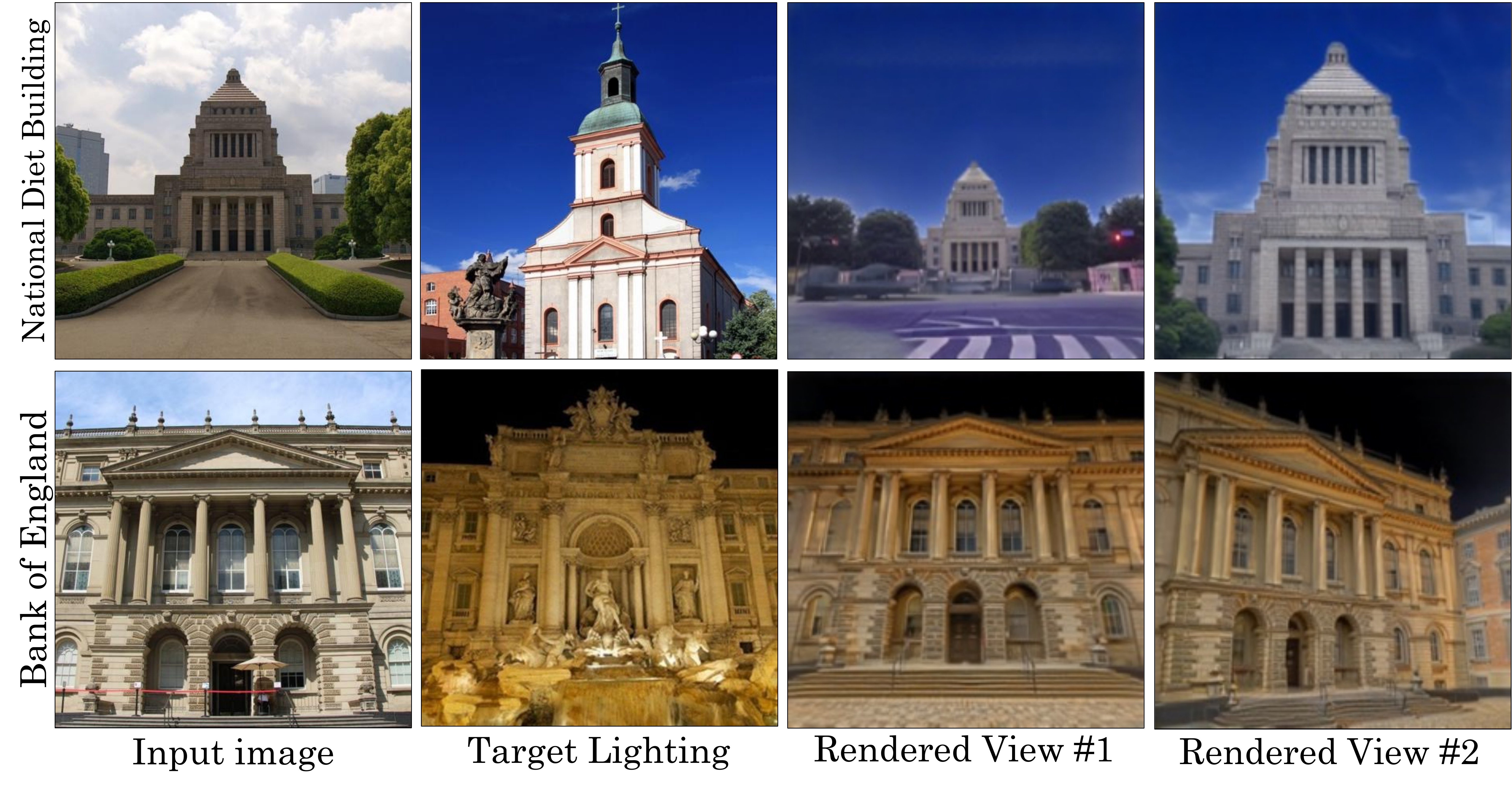}
    \vspace{-7mm}
    \caption{\textbf{Cross-scene appearance transfer.} Our method disentangles appearance from geometry, allowing adaptation of illumination from \textit{different} scenes, something prior methods~\cite{kulhanek2024wildgaussians, tang2024nexussplats} cannot do as they jointly optimize view and appearance.} 
    \vspace{-4mm}
    \label{fig:difflight}
\end{figure}

For each input image, the network predicts scene geometry (per-Gaussian parameters and depth), while the light encoder extracts a compact light code that represents the image's illumination. 
The appearance adapter conditions on this code and the canonical Gaussian colors to produce transformed colors, which are rasterized and compared to the original image. 
Though the method is not trained to render novel views or lighting, our method generalizes well to unseen views (Fig.~\ref{fig:megascenes-sota}). 
The network learns stable lighting representations that enable transferring illumination appearance from one scene to another (Fig.~\ref{fig:difflight}), a capability absent in prior in-the-wild methods~\cite{zhang2024gaussian, kulhanek2024wildgaussians, tang2024nexussplats}.

\section{Experiments}

\begin{figure*}
    \centering
    \includegraphics[width=\linewidth]{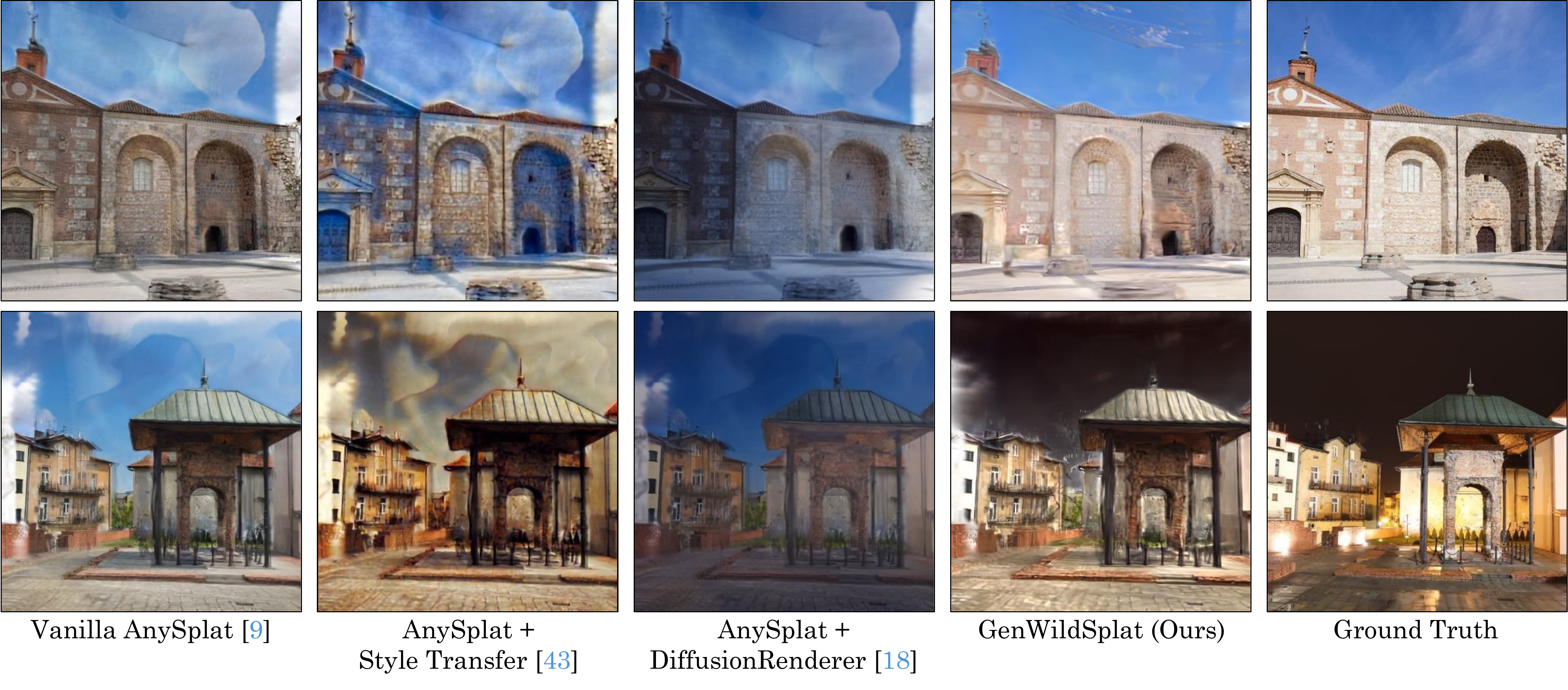}
    \vspace{-8mm}
\caption{\textbf{Comparison on the MegaScenes dataset against feed-forward based methods.} 
Existing feed-forward 3D Gaussian Splatting methods cannot handle unconstrained inputs, so we construct baselines using style transfer and DiffusionRenderer to address appearance variations. The DiffusionRenderer+AnySplat baseline integrates AnySplat with DiffusionRenderer, which uses environment maps from DiffusionLight-Turbo. Style transfer~\cite{wu2022ccpl} often introduces artifacts and color bleeding, while DiffusionRenderer~\cite{liang2025diffusion} produces unrealistic outdoor relighting (row 2 shows a dimmed, non-photorealistic “night”). These per-image methods suffer from multi-view inconsistency, whereas GenWildSplat modulates appearance in 3D, yielding photorealistic, view-consistent results.
}
    \vspace{-5mm}
    \label{fig:megascenes-baselines}
\end{figure*}

\subsection{Implementation Details}

GenWildSplat uses a 24-layer transformer with alternating frame and global attention. The depth, camera, and Gaussian heads adopt a DPT-based architecture that fuses multi-scale features to predict per-pixel depth, Gaussian attributes, and camera parameters. The light encoder follows~\cite{zhang2024latent}, producing 16-dimensional lighting vectors. An MLP expands these to 75 dimensions and modulates the per-Gaussian SH coefficients accordingly. For occlusion detection, we use YOLOv8 Segmentation~\cite{yolov8_ultralytics} to classify common COCO categories (person, car, dog, etc.) and merge them into a binary transient mask. The model, initialized from AnySplat pre-trained weights, uses a perceptual loss weight of $\lambda=0.05$ and is trained via curriculum learning for 40K iterations (Stage 1: 10K, Stage 2: 10K, Stage 3: 20K) over 2 days on a single RTX A6000. Since this problem setting is highly challenging, we additionally apply SyncFix~\cite{li2026syncfix} as a post-processing step to enhance the final results. We use this step only for visualization purposes and exclude it from all baseline comparisons, so that differences between methods remain clearly visible. All figures and videos showing only our method in this paper and on the project website are post-processed with SyncFix. Please refer to supplementary for more details. 

\subsection{Datasets}

\textbf{Training.} We train on 700+ outdoor scenes from DL3DV~\cite{ling2024dl3dv}, augmented with synthetic lighting and occlusions to mimic in-the-wild variability. Illumination diversity is produced using DiffusionRenderer~\cite{liang2025diffusion} via offline unconditioned relighting (~30 minutes per scene). Transient occluders are generated by compositing COCO segments~\cite{lin2014microsoft} (e.g., people, cars) at random locations, providing exact ground-truth masks and are applied on-the-fly.

\noindent \textbf{Evaluation.} We evaluate on PhotoTourism~\cite{snavely2006photo} using 6 input views across 3 scenes. To assess generalization, we further curate 20 challenging MegaScenes with strong lighting variation, occlusions, and viewpoint sparsity, selecting scenes with fewer than 20 registered images. This avoids artificial subsampling and provides a more realistic sparse-view benchmark for the future. Refer to the supplementary for visualizations of these sparse-view scenes.

\subsection{Baselines}
Recent in-the-wild reconstruction methods, such as GS-W~\cite{zhang2024gaussian}, WildGaussians~\cite{kulhanek2024wildgaussians}, and NexusSplats~\cite{tang2024nexussplats}, require per-scene and test-time optimization, making a direct comparison with our feed-forward approach infeasible. Methods like SparseGS-W~\cite{li2025sparsegs} and MS-GS~\cite{li2025ms} lack public implementations. We therefore define feed-forward baselines:

\noindent \textbf{AnySplat}~\cite{jiang2025anysplat} provides real-time reconstruction but does not model appearance variation.

\noindent \textbf{StyleTransfer-AnySplat} extends AnySplat with CCPL~\cite{wu2022ccpl} for style adaptation, though this adjusts artistic style rather than realistic illumination.

\noindent \textbf{DiffusionRenderer+AnySplat} integrates AnySplat with DiffusionRenderer~\cite{liang2025diffusion}, which models lighting using environment maps from DiffusionLight-Turbo~\cite{chinchuthakun2025diffusionlight}.

All baselines use Stable Diffusion~\cite{rombach2021highresolution} for mask-based inpainting to handle occlusions. 
We use AnySplat as our primary baseline, since GenWildSplat builds upon it; however, our modular approach could also extend to other feed-forward methods, such as MVSplat~\cite{chen2024mvsplat} or PixelSplat~\cite{charatan2024pixelsplat}.

\begin{table}[t]
\centering
\caption{\textbf{Quantitative comparison against baselines.} We compare against in-the-wild baselines on MegaScenes under sparse-view settings with varying input views, where the \colorbox{red!40}{best} scores and \colorbox{orange!50}{second best} scores are highlighted with respective colors.}
\vspace{-2mm}
\label{tab:phototourism_megascenes_baselines}
\setlength{\tabcolsep}{5pt}
\renewcommand{\arraystretch}{1.1}
\resizebox{0.45\textwidth}{!}{
\begin{tabular}{l c c ccc c ccc}
\toprule
\multirow{2}{*}{\textbf{Method}} & \multirow{2}{*}{\textbf{Gen.}} & \multirow{2}{*}{\textbf{Time}} &
\multicolumn{3}{c}{\textbf{MegaScenes (3-View)}} & &
\multicolumn{3}{c}{\textbf{MegaScenes (6-View)}} \\
\cmidrule(lr){4-6} \cmidrule(lr){8-10}
 & & & \textbf{PSNR} & \textbf{SSIM} & \textbf{LPIPS} & &
 \textbf{PSNR} & \textbf{SSIM} & \textbf{LPIPS} \\
\midrule
GS-W & \xmark & 5 hrs & {11.60} & {0.285} & {0.623} & & {12.01} & {0.312} & {0.552} \\
WildGaussians & \xmark & 8 hrs & {12.73} & {0.316} & {0.599} & & {13.29} & {0.373} & {0.532} \\
NexusSplats & \xmark & 2.4 hrs & \ps{13.17} & \ps{0.335} & \ps{0.552} & & \ps{13.92} & \ps{0.397} & \ps{0.518} \\
\midrule
\textbf{GenWildSplat (Ours)} & \cmark & 3 secs & \pf{14.43} & \pf{0.402} & \pf{0.496} & & \pf{15.84} & \pf{0.440} & \pf{0.407} \\
\bottomrule
\vspace{-7mm}
\end{tabular}
}
\end{table}

\begin{table}[t]
\centering
\caption{\textbf{Quantitative comparison against feed-forward methods.} We compare our method against the feed-forward baselines on the sparse-view setting on the MegaScenes dataset. }
\vspace{-2mm}
\label{tab:megascenes_anysplat}
\setlength{\tabcolsep}{6pt}
\renewcommand{\arraystretch}{1.1}
\resizebox{0.45\textwidth}{!}{
\begin{tabular}{l c c c c}
\toprule
\textbf{Method} & \textbf{View Consistent} & \textbf{PSNR} & \textbf{SSIM} & \textbf{LPIPS} \\
\midrule
Vanilla AnySplat & \xmark & 12.65 & \ps{0.311} & \ps{0.412} \\
2D Baseline + AnySplat & \xmark & 12.90 & 0.281 & 0.486 \\
% MV Baseline + AnySplat & \xmark & -- & -- & -- \\
\midrule
DiffusionRenderer + AnySplat & \xmark & \ps{13.59} & 0.309 & 0.444 \\
% DiLightNet + Inpaint + AnySplat & \cmark & -- & -- & -- \\
% RGB$\leftrightarrow$X + Inpaint + AnySplat & \cmark & -- & -- & -- \\
\midrule
\textbf{GenWildSplat (Ours)} & \cmark & \pf{15.84} & \pf{0.440} & \pf{0.407} \\
\bottomrule
\end{tabular}
}
\end{table}

\subsection{Comparison on the PhotoTourism Dataset}
We evaluate GenWildSplat against state-of-the-art in-the-wild baselines~\cite{zhang2024gaussian, kulhanek2024wildgaussians, tang2024nexussplats} on sparse-view PhotoTourism (Fig.~\ref{fig:phototourism-sota}, Tab.~\ref{tab:phototourism_megascenes_baselines}). As Fig.~\ref{fig:failure} shows, COLMAP poses often degrade baseline performance under sparse views, so we use VGGT poses for fair comparison. Despite no scene-specific training, our feed-forward model surpasses optimization-based methods, producing more realistic renderings from sparse inputs. This stems from our appearance adapter, which transfers priors learned via curriculum training, enabling inference in just 3 seconds.

\subsection{Comparison on the MegaScenes Dataset}
We further evaluate GenWildSplat on the challenging MegaScenes dataset (Fig.~\ref{fig:megascenes-sota}, Tab.~\ref{tab:megascenes_anysplat}). Prior methods, trained from scratch without learned priors, exhibit severe artifacts and distortions, such as noisy ground regions(row 1), poor generalisation to novel views(row 2), and resulting in blurred skies(row 3). GenWildSplat produces clean, consistent renderings across diverse scenes, demonstrating its robustness even on very challenging in-the-wild datasets.

For a fair comparison, we benchmark against a few AnySplat variants (Fig.~\ref{fig:megascenes-baselines}). 
2D style transfer~\cite{wu2022ccpl} often introduces artifacts and color bleeding, while DiffusionRenderer~\cite{liang2025diffusion}, relying on estimated environment maps, produces unrealistic outdoor relighting (e.g., row 2 shows a dimmed but non-photorealistic “night”). These per-image methods suffer from multi-view inconsistency, unlike GenWildSplat, which modulates appearance directly in 3D for photorealistic, view-consistent results. 

\subsection{Results with Lighting from Different Scene} 
Unlike prior in-the-wild methods that jointly optimize lighting and geometry and require a target lighting image from the same scene, GenWildSplat disentangles appearance from geometry, enabling cross-scene illumination transfer. As shown in Fig.~\ref{fig:difflight}, it produces photorealistic, view-consistent results while preserving spatial and structural consistency, demonstrating robust appearance control.

\begin{table}[t]
\centering
\caption{\textbf{Ablation study} evaluated on the MegaScenes Dataset.}
\vspace{-2mm}
\resizebox{0.8\linewidth}{!}{
\begin{tabular}{lccc}
\toprule
\textbf{Model Variant} & \textbf{PSNR} $\uparrow$ & \textbf{SSIM} $\uparrow$ & \textbf{LPIPS} $\downarrow$ \\
\midrule
w/o Appearance adapter & 13.76 & 0.391 & \pf{0.405} \\
w/o Occlusion handling & \ps{15.14} & \ps{0.405} & 0.513 \\
w/o Curriculum learning & 11.72 & 0.318 & 0.438 \\
\textbf{Full model (Ours)} & \pf{15.84} & \pf{0.440} & \ps{0.407}  \\
\bottomrule

\end{tabular}}
\label{tab:ablation_components}
\vspace{-1mm}
\end{table}

\begin{figure}
    \centering
    \includegraphics[width=\linewidth]{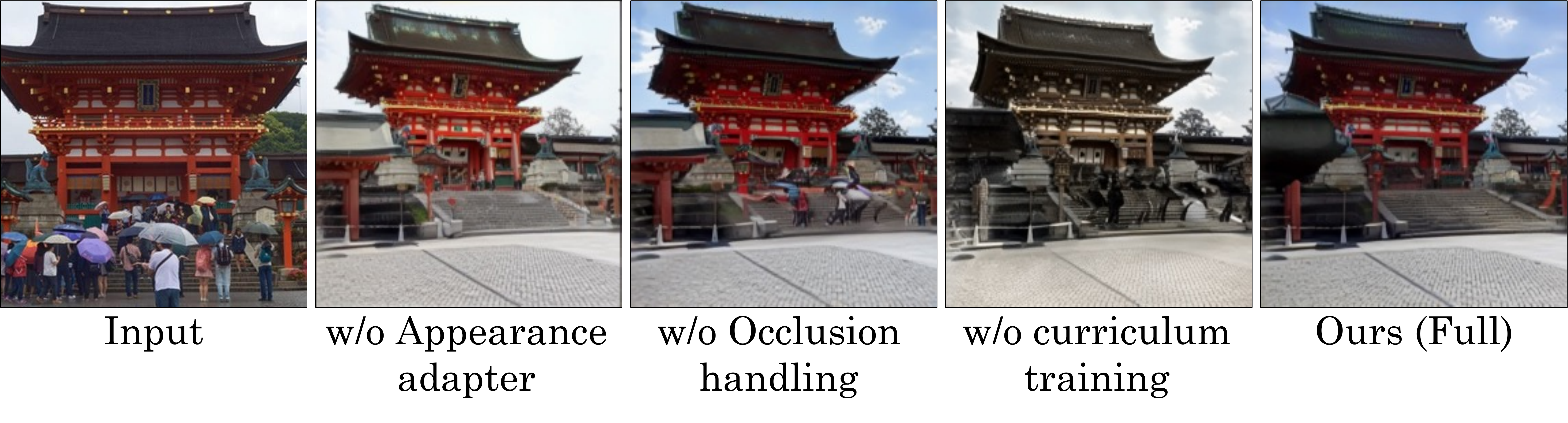}
    \vspace{-7mm}
    \caption{\textbf{Ablation Study.} Removing the appearance adapter, occlusion handling, or curriculum causes major failures: fixed appearance, baked-in transient objects, or color collapse. With all components enabled, GenWildSplat produces clean, consistent 3D reconstructions.}
    \vspace{-4mm}
    \label{fig:ablation}
\end{figure}

\subsection{Ablation Study \& Analysis}
To evaluate the contribution of each component in our method, we perform an ablation study shown in Fig.~\ref{fig:ablation} and Tab.~\ref{tab:ablation_components}. Removing the appearance adapter prevents the model from capturing appearance variations, resulting in a fixed, single appearance. Disabling occlusion handling prevents the removal of transient objects, such as people on the stairs.
Without the proposed curriculum-based training, the Gaussian colors collapse, as the model struggles to learn geometry, appearance, and occlusions simultaneously. 
With all components enabled, GenWildSplat models both appearance and occlusions, producing view-consistent renderings.

\begin{figure}
    \centering
    \includegraphics[width=\linewidth]{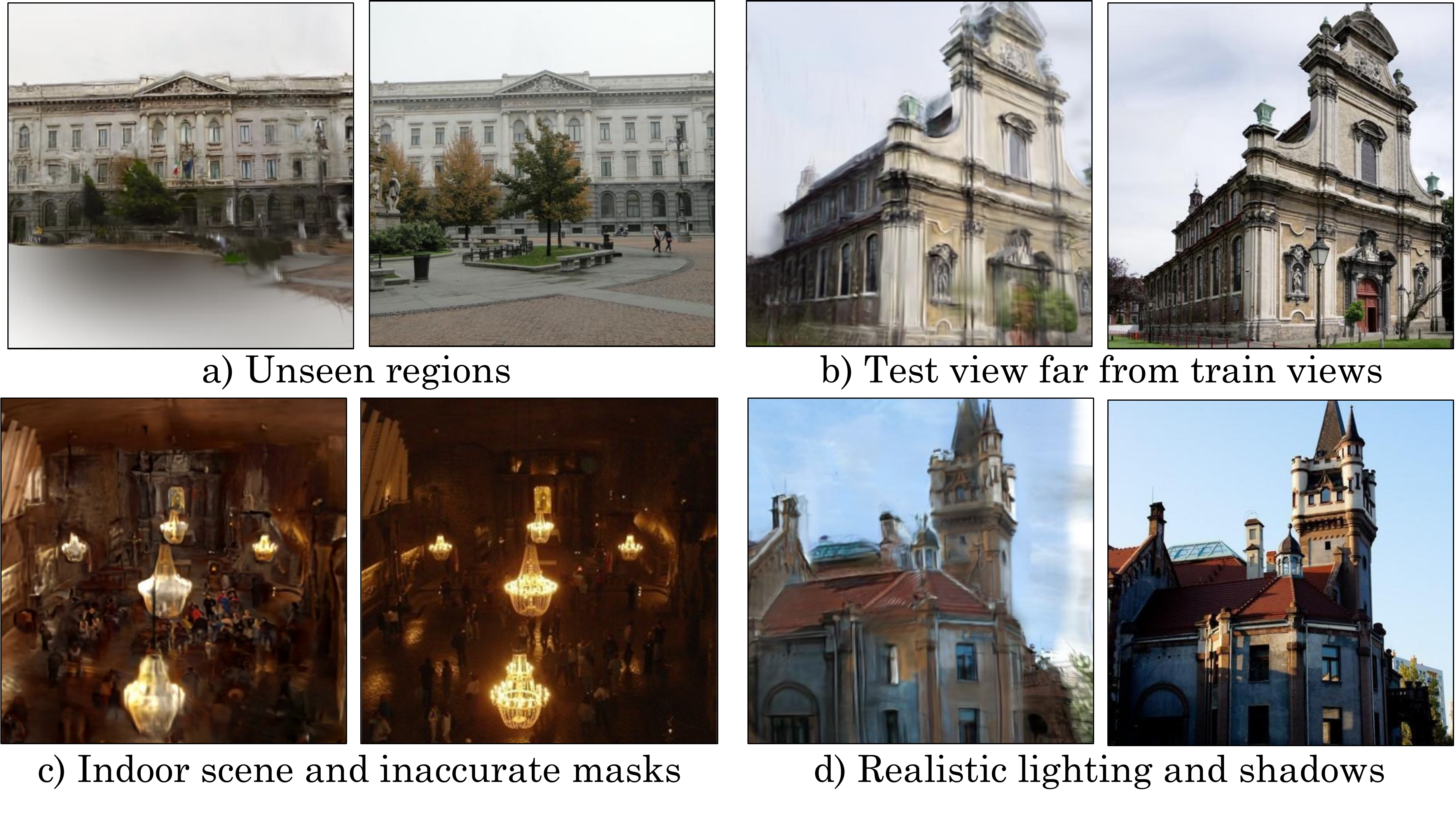}
    \vspace{-7mm}
    \caption{\textbf{Limitations.} (a) missing geometry in sparsely observed regions, (b) artifacts and double geometry for test views distant from training views, (c) degraded performance in indoor environments with imperfect occlusion masks, and (d) absence of shadow modeling and realistic relighting.}
    \vspace{-6mm}
    \label{fig:indoor}
\end{figure}

\section{Discussions}
% \noindent \textbf{Limitations.} GenWildSplat demonstrates strong feed-forward reconstruction under sparse, in-the-wild conditions, but several limitations remain. First, it relies on occlusion masks from a pre-trained segmentation model; while common transient classes are covered, rare or novel occluders (e.g., balloons, banners, flags) may be missed. This could be mitigated by training a multi-view segmentation network that detects inconsistencies across views and determines which objects should contribute to the final 3D reconstruction. Second, training is self-supervised to reproduce input views only; although this encourages implicit disentanglement of geometry and lighting, large viewpoint extrapolations can produce artifacts and doubled geometry. Adopting a strategy like RayZer~\cite{jiang2025rayzer} could address this issue. Third, our light encoder is a simple CNN originally designed for indoor scenes~\cite{zhang2024latent}, whereas outdoor illumination is more complex and may benefit from more expressive architectures, such as transformers. Finally, due to compute constraints, training was limited to a single GPU with ~700 synthetic scenes; scaling to larger synthetic or real-world datasets (e.g., MegaScenes) would likely improve robustness and fidelity.

\noindent \textbf{Limitations.}
GenWildSplat, while effective under sparse, in-the-wild image collections, has several limitations. First, sparse viewpoints naturally leave unseen regions, leading to incomplete geometry in areas not covered by the input images. 
Second, when test views lie far outside the training distribution, the model may produce artifacts or double-layered geometry due to limited viewpoint generalization. Third, indoor scenes are still hard: if the occlusion mask fails to accurately capture objects or depth discontinuities, the resulting masks degrade the reconstruction quality. 
Finally, the method does not model cast shadows or support realistic relighting, limiting its applicability to tasks that require physically consistent illumination.

\vspace{1mm}
\noindent \textbf{Conclusion.}
We present GenWildSplat, a generalizable, feed-forward Gaussian-splatting framework that reconstructs 3D scenes from sparse, unconstrained photo collections in under 3 seconds. 
The key to our success is the appearance adapter, which directly modulates Gaussian colors in 3D, and a robust occlusion handling mechanism, producing view-consistent, photorealistic renderings. GenWildSplat moves the needle towards real-time, controllable, relightable 3D scenes from sparse internet imagery.
%By enabling real-time controllable relighting and reliable occlusion management from sparse internet imagery, GenWildSplat paves the way for practical applications in AR/VR, virtual tourism, and navigation.

\vspace{1mm}
\noindent \textbf{Acknowledgements.} We are grateful for the valuable feedback and insightful discussions provided by Mukund Varma T, Yao-Chih Lee, Yu-Hsiang Huang, Hadi Alzayer, Sai Sri Teja Kuppa, and S Bala Prasanna. 
% \vspace{-6mm}
% \clearpage
{
    \small
    \bibliographystyle{ieeenat_fullname}
    \bibliography{main}

@String(CVPR= {IEEE Conf. Comput. Vis. Pattern Recog.})

@String(ICCV= {Int. Conf. Comput. Vis.})

@String(ECCV= {Eur. Conf. Comput. Vis.})

@String(ICLR = {Int. Conf. Learn. Represent.})

@String(CVPR  = {CVPR})

@String(ICCV  = {ICCV})

@String(ECCV  = {ECCV})

@String(ICLR  = {ICLR})

@article{kerbl20233d,
  title={3d gaussian splatting for real-time radiance field rendering.},
  author={Kerbl, Bernhard and Kopanas, Georgios and Leimk{\"u}hler, Thomas and Drettakis, George},
  journal={ACM Trans. Graph.},
  volume={42},
  number={4},
  pages={139--1},
  year={2023}
}

@inproceedings{xu2024wild,
  title={Wild-gs: Real-time novel view synthesis from unconstrained photo collections},
  author={Xu, Jiacong and Mei, Yiqun and Patel, Vishal},
  booktitle={NeurIPS},
  year={2024}
}

@inproceedings{ling2024dl3dv,
  title={Dl3dv-10k: A large-scale scene dataset for deep learning-based 3d vision},
  author={Ling, Lu and Sheng, Yichen and Tu, Zhi and Zhao, Wentian and Xin, Cheng and Wan, Kun and Yu, Lantao and Guo, Qianyu and Yu, Zixun and Lu, Yawen and others},
  booktitle={CVPR},
  year={2024}
}

@article{mur2015orb,
  title={ORB-SLAM: A versatile and accurate monocular SLAM system},
  author={Mur-Artal, Raul and Montiel, Jose Maria Martinez and Tardos, Juan D},
  journal={IEEE transactions on robotics},
  volume={31},
  number={5},
  pages={1147--1163},
  year={2015},
  publisher={IEEE}
}

@article{zhou2024comprehensive,
  title={A comprehensive review of vision-based 3d reconstruction methods},
  author={Zhou, Linglong and Wu, Guoxin and Zuo, Yunbo and Chen, Xuanyu and Hu, Hongle},
  journal={Sensors},
  volume={24},
  number={7},
  pages={2314},
  year={2024},
}

@inproceedings{lin2014microsoft,
  title={Microsoft coco: Common objects in context},
  author={Lin, Tsung-Yi and Maire, Michael and Belongie, Serge and Hays, James and Perona, Pietro and Ramanan, Deva and Doll{\'a}r, Piotr and Zitnick, C Lawrence},
  booktitle={ECCV},
  year={2014},
  
}

@inproceedings{chen2022hallucinated,
  title={Hallucinated neural radiance fields in the wild},
  author={Chen, Xingyu and Zhang, Qi and Li, Xiaoyu and Chen, Yue and Feng, Ying and Wang, Xuan and Wang, Jue},
  booktitle={CVPR},
  year={2022}
}

@inproceedings{tung2024megascenes,
  title={Megascenes: Scene-level view synthesis at scale},
  author={Tung, Joseph and Chou, Gene and Cai, Ruojin and Yang, Guandao and Zhang, Kai and Wetzstein, Gordon and Hariharan, Bharath and Snavely, Noah},
  booktitle={ECCV},
  year={2024},
  
}

@inproceedings{rudnev2022nerf,
  title={Nerf for outdoor scene relighting},
  author={Rudnev, Viktor and Elgharib, Mohamed and Smith, William and Liu, Lingjie and Golyanik, Vladislav and Theobalt, Christian},
  booktitle={ECCV},
  year={2022},
  
}

@inproceedings{sun2022neural,
  title={Neural 3d reconstruction in the wild},
  author={Sun, Jiaming and Chen, Xi and Wang, Qianqian and Li, Zhengqi and Averbuch-Elor, Hadar and Zhou, Xiaowei and Snavely, Noah},
  booktitle={ACM SIGGRAPH 2022 conference proceedings},
  year={2022}
}

@inproceedings{zhang2024gaussian,
  title={Gaussian in the wild: 3d gaussian splatting for unconstrained image collections},
  author={Zhang, Dongbin and Wang, Chuming and Wang, Weitao and Li, Peihao and Qin, Minghan and Wang, Haoqian},
  booktitle={ECCV},
  year={2024},
  
}

@inproceedings{dahmani2024swag,
  title={Swag: Splatting in the wild images with appearance-conditioned gaussians},
  author={Dahmani, Hiba and Bennehar, Moussab and Piasco, Nathan and Roldao, Luis and Tsishkou, Dzmitry},
  booktitle={ECCV},
  year={2024},
  
}

@article{yin2024fewviewgs,
  title={FewViewGS: Gaussian splatting with few view matching and multi-stage training},
  author={Yin, Ruihong and Yugay, Vladimir and Li, Yue and Karaoglu, Sezer and Gevers, Theo},
  journal={NeurIPS},
  year={2024}
}

@inproceedings{zhang2024latent,
  title={Latent intrinsics emerge from training to relight},
  author={Zhang, Xiao and Gao, William and Jain, Seemandhar and Maire, Michael and Forsyth, David and Bhattad, Anand},
  booktitle={NeurIPS},
  year={2024}
}

@inproceedings{sabour2023robustnerf,
  title={Robustnerf: Ignoring distractors with robust losses},
  author={Sabour, Sara and Vora, Suhani and Duckworth, Daniel and Krasin, Ivan and Fleet, David J and Tagliasacchi, Andrea},
  booktitle={CVPR},
  year={2023}
}

@inproceedings{ren2024nerf,
  title={Nerf on-the-go: Exploiting uncertainty for distractor-free nerfs in the wild},
  author={Ren, Weining and Zhu, Zihan and Sun, Boyang and Chen, Jiaqi and Pollefeys, Marc and Peng, Songyou},
  booktitle={CVPR},
  year={2024}
}

@article{wang2024we,
  title={We-gs: An in-the-wild efficient 3d gaussian representation for unconstrained photo collections},
  author={Wang, Yuze and Wang, Junyi and Qi, Yue},
  journal={arXiv preprint arXiv:2406.02407},
  year={2024}
}

@inproceedings{kulhanek2024wildgaussians,
  title={WildGaussians: 3D Gaussian splatting in the wild},
  author={Kulhanek, Jonas and Peng, Songyou and Kukelova, Zuzana and Pollefeys, Marc and Sattler, Torsten},
  booktitle={NeurIPS},
  year={2024}
}

@article{tang2024nexussplats,
  title={Nexussplats: Efficient 3d gaussian splatting in the wild},
  author={Tang, Yuzhou and Xu, Dejun and Hou, Yongjie and Wang, Zhenzhong and Jiang, Min},
  journal={arXiv preprint arXiv:2411.14514},
  year={2024}
}

@incollection{snavely2006photo,
  title={Photo tourism: exploring photo collections in 3D},
  author={Snavely, Noah and Seitz, Steven M and Szeliski, Richard},
  booktitle={ACM siggraph 2006 papers},
  year={2006}
}

@inproceedings{jiang2025anysplat,
  title={AnySplat: Feed-forward 3D Gaussian Splatting from Unconstrained Views},
  author={Jiang, Lihan and Mao, Yucheng and Xu, Linning and Lu, Tao and Ren, Kerui and Jin, Yichen and Xu, Xudong and Yu, Mulin and Pang, Jiangmiao and Zhao, Feng and others},
  booktitle={ACM SIGGRAPH Asia},
  year={2025}
}

@inproceedings{martin2021nerf,
  title={Nerf in the wild: Neural radiance fields for unconstrained photo collections},
  author={Martin-Brualla, Ricardo and Radwan, Noha and Sajjadi, Mehdi SM and Barron, Jonathan T and Dosovitskiy, Alexey and Duckworth, Daniel},
  booktitle={CVPR},
  year={2021}
}

@article{oquab2023dinov2,
  title={DINOv2: Learning Robust Visual Features without Supervision},
  author={Oquab, Maxime and Darcet, Timoth{\'e}e and Moutakanni, Th{\'e}o and Vo, Huy and Szafraniec, Marc and Khalidov, Vasil and Fernandez, Pierre and Haziza, Daniel and Massa, Francisco and El-Nouby, Alaaeldin and others},
  journal={Transactions on Machine Learning Research Journal},
  pages={1--31},
  year={2024}
}

@inproceedings{wang2025vggt,
  title={Vggt: Visual geometry grounded transformer},
  author={Wang, Jianyuan and Chen, Minghao and Karaev, Nikita and Vedaldi, Andrea and Rupprecht, Christian and Novotny, David},
  booktitle={CVPR},
  pages={5294--5306},
  year={2025}
}

@inproceedings{wang2024dust3r,
  title={Dust3r: Geometric 3d vision made easy},
  author={Wang, Shuzhe and Leroy, Vincent and Cabon, Yohann and Chidlovskii, Boris and Revaud, Jerome},
  booktitle={CVPR},
  year={2024}
}

@inproceedings{mast3r,
  title={Grounding image matching in 3d with mast3r},
  author={Leroy, Vincent and Cabon, Yohann and Revaud, J{\'e}r{\^o}me},
  booktitle={ECCV},
  year={2024},
  
}

@inproceedings{lrm,
  title={LRM: Large Reconstruction Model for Single Image to 3D},
  author={Hong, Yicong and Zhang, Kai and Gu, Jiuxiang and Bi, Sai and Zhou, Yang and Liu, Difan and Liu, Feng and Sunkavalli, Kalyan and Bui, Trung and Tan, Hao},
  booktitle={ICLR},
year={2023},
}

@inproceedings{gslrm,
  title={Gs-lrm: Large reconstruction model for 3d gaussian splatting},
  author={Zhang, Kai and Bi, Sai and Tan, Hao and Xiangli, Yuanbo and Zhao, Nanxuan and Sunkavalli, Kalyan and Xu, Zexiang},
  booktitle={ECCV},
  year={2024},
}

@inproceedings{chen2024mvsplat,
  title={Mvsplat: Efficient 3d gaussian splatting from sparse multi-view images},
  author={Chen, Yuedong and Xu, Haofei and Zheng, Chuanxia and Zhuang, Bohan and Pollefeys, Marc and Geiger, Andreas and Cham, Tat-Jen and Cai, Jianfei},
  booktitle={ECCV},
  year={2024},
}

@inproceedings{ziwen2024long,
  title={Long-lrm: Long-sequence large reconstruction model for wide-coverage gaussian splats},
  author={Ziwen, Chen and Tan, Hao and Zhang, Kai and Bi, Sai and Luan, Fujun and Hong, Yicong and Fuxin, Li and Xu, Zexiang},
  booktitle={ICCV},
  year={2025}
}

@inproceedings{jinlvsm,
  title={LVSM: A Large View Synthesis Model with Minimal 3D Inductive Bias},
  author={Jin, Haian and Jiang, Hanwen and Tan, Hao and Zhang, Kai and Bi, Sai and Zhang, Tianyuan and Luan, Fujun and Snavely, Noah and Xu, Zexiang},
  booktitle={ICLR},
year={2024}
}

@inproceedings{yang2025fast3r,
  title={Fast3r: Towards 3d reconstruction of 1000+ images in one forward pass},
  author={Yang, Jianing and Sax, Alexander and Liang, Kevin J and Henaff, Mikael and Tang, Hao and Cao, Ang and Chai, Joyce and Meier, Franziska and Feiszli, Matt},
  booktitle={CVPR},
  year={2025}
}

@inproceedings{zhu2024fsgs,
  title={Fsgs: Real-time few-shot view synthesis using gaussian splatting},
  author={Zhu, Zehao and Fan, Zhiwen and Jiang, Yifan and Wang, Zhangyang},
  booktitle={ECCV},
  year={2024},
  
}

@inproceedings{paliwal2024coherentgs,
  title={Coherentgs: Sparse novel view synthesis with coherent 3d gaussians},
  author={Paliwal, Avinash and Ye, Wei and Xiong, Jinhui and Kotovenko, Dmytro and Ranjan, Rakesh and Chandra, Vikas and Kalantari, Nima Khademi},
  booktitle={ECCV},
  year={2024},
  
}

@inproceedings{kaleta2025lumigauss,
  title={LumiGauss: Relightable Gaussian Splatting in the Wild},
  author={Kaleta, Joanna and Kania, Kacper and Trzci{\'n}ski, Tomasz and Kowalski, Marek},
  booktitle={WACV},
  year={2025},
  
}

@inproceedings{xu2025depthsplat,
  title={Depthsplat: Connecting gaussian splatting and depth},
  author={Xu, Haofei and Peng, Songyou and Wang, Fangjinhua and Blum, Hermann and Barath, Daniel and Geiger, Andreas and Pollefeys, Marc},
  booktitle={CVPR},
  year={2025}
}

@article{wang2025look,
  title={Look at the sky: Sky-aware efficient 3D Gaussian splatting in the wild},
  author={Wang, Yuze and Wang, Junyi and Gao, Ruicheng and Qu, Yansong and Duan, Wantong and Yang, Shuo and Qi, Yue},
  journal={IEEE Transactions on Visualization and Computer Graphics},
  year={2025},
  publisher={IEEE}
}

@inproceedings{xiong2025sparsegs,
  title={Sparsegs: Sparse view synthesis using 3d gaussian splatting},
  author={Xiong, Haolin and Muttukuru, Sairisheek and Xiao, Hanyuan and Upadhyay, Rishi and Chari, Pradyumna and Zhao, Yajie and Kadambi, Achuta},
  booktitle={2025 International Conference on 3D Vision (3DV)},
  year={2025},
  
}

@inproceedings{tang2025mv,
  title={Mv-dust3r+: Single-stage scene reconstruction from sparse views in 2 seconds},
  author={Tang, Zhenggang and Fan, Yuchen and Wang, Dilin and Xu, Hongyu and Ranjan, Rakesh and Schwing, Alexander and Yan, Zhicheng},
  booktitle={CVPR},
  year={2025}
}

@inproceedings{yang2023cross,
  title={Cross-ray neural radiance fields for novel-view synthesis from unconstrained image collections},
  author={Yang, Yifan and Zhang, Shuhai and Huang, Zixiong and Zhang, Yubing and Tan, Mingkui},
  booktitle={ICCV},
  year={2023}
}

@inproceedings{charatan2024pixelsplat,
  title={pixelsplat: 3d gaussian splats from image pairs for scalable generalizable 3d reconstruction},
  author={Charatan, David and Li, Sizhe Lester and Tagliasacchi, Andrea and Sitzmann, Vincent},
  booktitle={CVPR},
  year={2024}
}

@inproceedings{jiang2025rayzer,
  title={RayZer: A Self-supervised Large View Synthesis Model},
  author={Jiang, Hanwen and Tan, Hao and Wang, Peng and Jin, Haian and Zhao, Yue and Bi, Sai and Zhang, Kai and Luan, Fujun and Sunkavalli, Kalyan and Huang, Qixing and others},
  booktitle={ICCV},
  year={2025}
}

@article{chen2024mvsplat360,
  title={Mvsplat360: Feed-forward 360 scene synthesis from sparse views},
  author={Chen, Yuedong and Zheng, Chuanxia and Xu, Haofei and Zhuang, Bohan and Vedaldi, Andrea and Cham, Tat-Jen and Cai, Jianfei},
  journal={NeurIPS},
  year={2024}
}

@article{wang2024freesplat,
  title={Freesplat: Generalizable 3d gaussian splatting towards free view synthesis of indoor scenes},
  author={Wang, Yunsong and Huang, Tianxin and Chen, Hanlin and Lee, Gim Hee},
  journal={NeurIPS},
  volume={37},
  year={2024}
}

@inproceedings{wang2025continuous,
  title={Continuous 3d perception model with persistent state},
  author={Wang, Qianqian and Zhang, Yifei and Holynski, Aleksander and Efros, Alexei A and Kanazawa, Angjoo},
  booktitle={CVPR},
  year={2025}
}

@inproceedings{xu2024grm,
  title={Grm: Large gaussian reconstruction model for efficient 3d reconstruction and generation},
  author={Xu, Yinghao and Shi, Zifan and Yifan, Wang and Chen, Hansheng and Yang, Ceyuan and Peng, Sida and Shen, Yujun and Wetzstein, Gordon},
  booktitle={ECCV},
  year={2024},
  
}

@inproceedings{wang20243d,
  title={3d reconstruction with spatial memory},
  author={Wang, Hengyi and Agapito, Lourdes},
  booktitle={2025 International Conference on 3D Vision (3DV)},
  year={2025},
  
}

@inproceedings{liu2025slam3r,
  title={Slam3r: Real-time dense scene reconstruction from monocular rgb videos},
  author={Liu, Yuzheng and Dong, Siyan and Wang, Shuzhe and Yin, Yingda and Yang, Yanchao and Fan, Qingnan and Chen, Baoquan},
  booktitle={CVPR},
  year={2025}
}

@inproceedings{murai2025mast3r,
  title={MASt3R-SLAM: Real-time dense SLAM with 3D reconstruction priors},
  author={Murai, Riku and Dexheimer, Eric and Davison, Andrew J},
  booktitle={CVPR},
  year={2025}
}

@software{yolov8_ultralytics,
  author = {Glenn Jocher and Ayush Chaurasia and Jing Qiu},
  title = {Ultralytics YOLOv8},
  version = {8.0.0},
  year = {2023},
  url = {https://github.com/ultralytics/ultralytics},
  license = {AGPL-3.0}
}

@article{li2025sparsegs,
  title={SparseGS-W: Sparse-View 3D Gaussian Splatting in the Wild with Generative Priors},
  author={Li, Yiqing and Wang, Xuan and Wu, Jiawei and Ma, Yikun and Jin, Zhi},
  journal={arXiv preprint arXiv:2503.19452},
  year={2025}
}

@article{li2025ms,
  title={MS-GS: Multi-Appearance Sparse-View 3D Gaussian Splatting in the Wild},
  author={Li, Deming and Jiang, Kaiwen and Tang, Yutao and Ramamoorthi, Ravi and Chellappa, Rama and Peng, Cheng},
  journal={arXiv preprint arXiv:2509.15548},
  year={2025}
}

@inproceedings{wu2022ccpl,
  title={Ccpl: Contrastive coherence preserving loss for versatile style transfer},
  author={Wu, Zijie and Zhu, Zhen and Du, Junping and Bai, Xiang},
  booktitle={ECCV},
  year={2022},
  
}

@inproceedings{liang2025diffusion,
  title={Diffusion Renderer: Neural Inverse and Forward Rendering with Video Diffusion Models},
  author={Liang, Ruofan and Gojcic, Zan and Ling, Huan and Munkberg, Jacob and Hasselgren, Jon and Lin, Chih-Hao and Gao, Jun and Keller, Alexander and Vijaykumar, Nandita and Fidler, Sanja and others},
  booktitle={CVPR},
  year={2025}
}

@article{chinchuthakun2025diffusionlight,
  title={DiffusionLight-Turbo: Accelerated Light Probes for Free via Single-Pass Chrome Ball Inpainting},
  author={Chinchuthakun, Worameth and Phongthawee, Pakkapon and Raj, Amit and Jampani, Varun and Khungurn, Pramook and Suwajanakorn, Supasorn},
  journal={arXiv preprint arXiv:2507.01305},
  year={2025}
}

@inproceedings{rombach2021highresolution,
  title={High-resolution image synthesis with latent diffusion models},
  author={Rombach, Robin and Blattmann, Andreas and Lorenz, Dominik and Esser, Patrick and Ommer, Bj{\"o}rn},
  booktitle={CVPR},
  year={2022}
}

@inproceedings{alzayer2025generative,
  title={Generative multiview relighting for 3D reconstruction under extreme illumination variation},
  author={Alzayer, Hadi and Henzler, Philipp and Barron, Jonathan T and Huang, Jia-Bin and Srinivasan, Pratul P and Verbin, Dor},
  booktitle={Proceedings of the Computer Vision and Pattern Recognition Conference},
  pages={10933--10942},
  year={2025}
}

@article{alzayer2025coupled,
  title={Coupled Diffusion Sampling for Training-Free Multi-View Image Editing},
  author={Alzayer, Hadi and Zhang, Yunzhi and Geng, Chen and Huang, Jia-Bin and Wu, Jiajun},
  journal={arXiv preprint arXiv:2510.14981},
  year={2025}
}

@article{li2026syncfix,
  title={SyncFix: Fixing 3D Reconstructions via Multi-View Synchronization},
  author={Li, Deming and Yadav, Abhay and Peng, Cheng and Chellappa, Rama and Bhattad, Anand},
  journal={arXiv preprint arXiv:2604.11797},
  year={2026}
}
}

\clearpage
\setcounter{page}{1}
\setcounter{section}{0}  % Reset section counter to 0 so next section = A

\renewcommand{\thesection}{\Alph{section}}
\renewcommand{\thesubsection}{\thesection.\arabic{subsection}}
\renewcommand{\thesubsubsection}{\thesubsection.\arabic{subsubsection}}

% "Appendix" heading without a letter label
\section*{Appendix}
\addcontentsline{toc}{section}{Appendix}  % optional: adds it to TOC

\section{Dataset Details}

\subsection{Training Dataset}
For training GenWildSplat, we constructed a large-scale synthetic dataset derived from the DL3DV~\cite{ling2024dl3dv} dataset. Initially, we subsampled 2,000 scenes from DL3DV, focusing specifically on outdoor environments, which resulted in approximately 700 scenes suitable for our goal. These scenes were then processed using our synthetic data generation pipeline to enhance appearance diversity and robustness. In particular, we employed DiffusionRenderer~\cite{liang2025diffusion} in a classifier-free guidance setting to randomly relight each image, thereby producing a wide range of lighting conditions. To optimize computational efficiency, we executed the inverse rendering step for only one iteration, as preliminary experiments indicated negligible differences compared to performing 15 iterations. The forward rendering step, however, was carried out for 15 iterations to ensure high-fidelity reconstructions of relit appearances. Due to the substantial computational cost of this process, approximately 30–45 minutes per scene, the relighting procedure was applied offline and limited to the 700 selected outdoor scenes.

We incorporated synthetic occlusions during training to mimic the in-the-wild images. Using a pretrained segmentation model~\cite{yolov8_ultralytics} on the COCO~\cite{lin2014microsoft} dataset, we created a comprehensive bank of objects that could serve as occluders. During training, we randomly sampled between 2 to 10 objects from this bank and positioned them in the lower half of the image, mimicing the empirical distribution of occlusions in real-world scenes. Occlusions were added on-the-fly, with corresponding occlusion masks generated in real time and used to supervise the model. This combination of relighting and occlusion augmentation enabled GenWildSplat to learn robust appearance and geometry representations under sparse inputs, diverse illumination, and realistic occlusions, ensuring strong generalization to unseen outdoor scenes.

\subsection{Evaluation Dataset}
For evaluation, we carefully curated a set of testing scenes from the MegaScenes dataset to reflect realistic sparsity and complexity. Specifically, we selected scenes containing fewer than 20–25 images, deliberately ensuring that the dataset mimics the sparse viewpoint coverage and diverse lighting conditions encountered in real-world captures. Unlike prior works such as MS-GS~\cite{li2025ms} and SparseGS-W~\cite{li2025sparsegs}, which simulate sparsity by artificially discarding images from densely captured scenes like PhotoTourism~\cite{snavely2006photo}, our selection prioritizes authenticity. By using scenes that are naturally sparse, we ensure that the evaluation closely represents practical scenarios where acquiring dense multi-view captures is infeasible.

Furthermore, the chosen scenes exhibit a range of illumination variations, including both subtle and extreme lighting changes, as well as moderate to high levels of transient occlusions. These characteristics create a challenging environment for novel-view synthesis and relighting tasks, providing a rigorous benchmark for assessing the performance and generalization capability of GenWildSplat and in total, we curated 20 scenes. 

\section{Additional Architecture Details}

\subsection{DPT Backbone}  
We adopt a Dense Prediction Transformer (DPT) backbone for predicting depth, camera parameters, and Gaussian scene representations. The DPT encoder generates multi-resolution feature maps that feed into three task-specific heads: (i) a depth head producing a dense depth map via convolutional fusion; (ii) a camera head estimating global pose and intrinsics using pooled high-level features followed by an MLP; and (iii) a Gaussian head that outputs per-Gaussian parameters (mean positions, anisotropic covariances, and feature vectors). This separation enables accurate spatial predictions while capturing global camera information in a compact latent representation.

\subsection{Light Encoder}  
The light encoder uses only the encoder portion of a U-Net style autoencoder built from residual convolutional blocks. Each block contains two convolutional layers, group normalization, and a nonlinear activation. The encoder has six resolution levels with block counts $[1,2,2,4,4,4]$, starting at $256\times256$ resolution and halving at each level. Latent channel widths are $[32,64,128,128,256,512]$. Extrinsic lighting features are extracted from the bottleneck using multiple MLP layers followed by spatial averaging, producing a 16-dimensional vector that captures low-frequency, global lighting. No intrinsic features are used.

\subsection{Segmentation}  
Occlusion masks are generated using the YOLOv8x.seg model trained on COCO classes. The selected COCO objects: person, bicycle, car, motorcycle, bus, train, truck, boat, bird, cat, dog, horse, sheep, cow, elephant, bear, zebra, giraffe, backpack, umbrella, handbag, suitcase, chair, keyboard, book. These objects are used to create an occluder bank for online augmentation during training.

\subsection{Appearance Adapter}  
The Appearance Adapter is a five-layer MLP mapping the concatenated 16-dimensional extrinsic light code and 75-dimensional per-Gaussian conditioning vector to 75 spherical-harmonic (SH) coefficients. Hidden layer sizes are $[256,512,512,256]$, with nonlinear activations after each layer and a linear output layer. This structure allows smooth low-frequency appearance modulation while retaining sufficient capacity to predict per-Gaussian SH lighting coefficients for rendering.

\end{document}